\documentclass[10pt,twocolumn,letterpaper]{article}

\usepackage{cvpr}
\usepackage{times}
\usepackage{graphicx}
\usepackage{amsmath}
\usepackage{amssymb}
\usepackage[american]{babel}
\usepackage{microtype}
\usepackage{multirow}
\usepackage[pagebackref=true,breaklinks=true,colorlinks,bookmarks=false]{hyperref}

\cvprfinalcopy % *** Uncomment this line for the final submission

\ifcvprfinal\fi
\begin{document}

\title{An Entropy-based Pruning Method for CNN Compression}

\author{Jian-Hao Luo \qquad Jianxin Wu\\
National Key Laboratory for Novel Software Technology \\
Nanjing University, China\\
{\tt\small \{luojh, wujx\}@lamda.nju.edu.cn}
}

\maketitle

\begin{abstract}
This paper aims to simultaneously accelerate and compress off-the-shelf CNN models via filter pruning strategy. The importance of each filter is evaluated by the proposed entropy-based method first. Then several unimportant filters are discarded to get a smaller CNN model. Finally, fine-tuning is adopted to recover its generalization ability which is damaged during filter pruning. Our method can reduce the size of intermediate activations, which would dominate most memory footprint during model training stage but is less concerned in previous compression methods. Experiments on the ILSVRC-12 benchmark demonstrate the effectiveness of our method. Compared with previous filter importance evaluation criteria, our entropy-based method obtains better performance. We achieve $3.3\times$ speed-up and $16.64\times$ compression on VGG-16, $1.54\times$ acceleration and $1.47\times$ compression on ResNet-50, both with about $1\%$ top-5 accuracy decrease.
\end{abstract}

\section{Introduction}

In the past few years, we have witnessed a rapid development of deep neural networks in the field of computer vision, from the basic image classification task such as the ImageNet recognition challenge~\cite{Krizhevsky12NIPS,Simonyan14ICLR,He16CVPR}, to some more advanced applications, e.g., object detection~\cite{Girshick14CVPR}, semantic segmentation~\cite{Long15CVPR}, image captioning~\cite{Johnson16CVPR} and many others. Deep neural networks have achieved state-of-the-art performance in these fields compared with traditional methods based on manually designed visual features. 

In spite of its great success, a typical deep model is hard to be deployed on resource constrained devices, \eg, mobile phones and embedded gadgets. A resource constrained scenario means a computing task must be accomplished with limited resource supply, such as computing time, storage space, battery power, computing capability and so on. One of the main issues of deep neural networks is its huge computational and storage overhead, which constitutes a serious challenge for a mobile device with limited computing resource~\cite{Han16ICLR}. For instance, the VGG-16 model~\cite{Simonyan14ICLR} has 138.34 million parameters, taking up more than 500MB storage space, and needs 15.5 billion float point operations (FLOPs) to classify a single image. Such a cumbersome model can easily exceed the computing limit of most small devices like cellphones. Thus, network compression has drawn a significant amount of interests from both academia and industry.

The resource consumption of CNN models mainly comes from three aspects: running time, model parameter size and intermediate activation size. In order to deploy CNN models on these small devices, numerous efficient compression methods have been proposed in the literature~\cite{Wu16CVPR, Rastegari16ECCV, Denton14NIPS, Han16ICLR, Hinton14NIPS}. Early works~\cite{Chen15ICML, Sindhwani15NIPS} mainly focus on compressing parameters of the fully-connected layers, which are suffering from great redundancy and are easy to compress. In order to accelerate model inference speed, compressing convolutional layers~\cite{Denton14NIPS, Wu16CVPR} becomes an important research direction too. But, almost none of them is concerned about reducing the size of intermediate activations. 

However, there is a great need for reducing intermediate activations. First, intermediate activations would dominate most memory footprint when batch size is large. For example, the intermediate activations of all layers in the VGG-16 model~\cite{Simonyan14ICLR} would take up almost 109.89MB memory space for \emph{one} image. Even in a forward inference stage, it would take up no less than 12.25MB. Such a huge memory consumption would quickly exceed the memory limit of most small devices with a \emph{large batch size}. In fact, in small devices it is impossible to fine-tune CNN models that are compressed with most current methods, due to the huge size of activations. Secondly, reducing activations would simultaneously accelerate inference speed and reduce memory footprint. Last but not least, the number of channels in each layer is manually designed during the training purpose. The 80-20 rule suggests that many filters will be redundant, \ie, having very little contribution to the overall performance.

In this paper, we propose an entropy-based framework to prune several unimportant filters to simultaneously accelerate and compress CNN models in both training and test stage, converting a cumbersome network into a much smaller model with minor performance degradation. Our main insight is that, if the activation output of a specific filter contains less information, this filter seems less important, thus should be removed to get a small network. We propose an entropy-based channel selection metric to evaluate the importance of each filter, and prune several weak filters. Then the pruned model is fine-tuned to regain its discrimination ability. Since the naive iterative pruning strategy is too time-consuming, we propose a learning schedule to seek a trade-off between training speed and model accuracy. 

We evaluate our pruning framework for image classification using two commonly used CNN models: VGG-16~\cite{Simonyan14ICLR} and the recently proposed ResNet-50~\cite{He16CVPR}. These two models are both pruned on the benchmark dataset ILSVRC-12 of the ImageNet repository~\cite{ILSVRC15}.\footnote{In this paper, we use ImageNet to denote the ILSVRC-12 dataset for simplicity. ILSVRC-12 is a commonly used subset of ImageNet.}  Our method achieves $3.3\times$ acceleration and $16.64\times$ compression on VGG-16, with about $1\%$ top-5 accuracy drop. As for ResNet-50, there exists less redundancy compared with classic CNN models. We can still bring $1.54\times$ acceleration and $1.47\times$ compression with roughly $1\%$ top-5 accuracy drop.

These acceleration and compression results are better than or comparable to existing methods~\cite{Han16ICLR, Wu16CVPR, Han15NIPS}, yet it is not our main consideration. Our goal is to convert a cumbersome network into a slim model \emph{that can run on off-the-shelf deep learning libraries without any modification}. We believe fine-tuning on new datasets or tasks is necessary (rather than only using the compressed network for ImageNet classification), but any changes on network layers and architecture would not be supported by current deep learning libraries. Thus, we believe pruning filters is the suitable strategy. We also compare our pruned network with the original model on two different domain-specific datasets, CUB-200-2011~\cite{CUB_200_2011} and Indoor-67~\cite{Indoor67}, to demonstrate the domain adaptation ability of our method. The major advantages and contributions are summarized as follows:
\begin{itemize}
	\item A simple yet effective framework is proposed, to accelerate and compress CNN models in both training and inference stage. Our method can compress the size of intermediate activations, reducing the run-time memory consumption dramatically, which is less concerned in previous works.
	\item We propose an effective learning schedule strategy to seek a better trade-off between training speed and classification accuracy in our iterative pruning framework.
%	\item The pruned network can be fine-tuned on other domain specific datasets, showing a good generalization ability.
	\item Our method does not rely on dedicated libraries to gain the acceleration and compression performance, thus could be perfectly supported by any current popular deep learning library.
\end{itemize}

\section{Related work}
Many researchers have revealed that the deep model is suffering from over-parameterization heavily. For example, Denil \etal~\cite{Denil13NIPS} demonstrated that a network can be efficiently reconstructed with only a small subset of its original parameters. What should be emphasized is that, this redundancy seems necessary during the model training stage, since the highly non-convex optimization problem is hard to be solved with current technique~\cite{Denton14NIPS, Hinton12arxiv}. Therefore, most compression strategies aim to convert a pre-trained model into a small scale model. In this section, we will give a brief introduction of these popular methods from several different aspects.

\textbf{Low-rank approximation:} In most deep models, the parameters of each layer form a large and dense matrix, which leads to both storage and computational difficulties. Mathematically, if we can approximate this dense matrix with several low-rank small scale matrices, the matrix-vector multiplication can be quickly finished with some special technique like the fast Fourier transform (FFT). Thus, both memory footprint and computational complexity can be reduced dramatically. Inspired by this idea, many methods have been proposed to approximate the original weights. Sindhwani \etal~\cite{Sindhwani15NIPS} utilize a linear combination of several structured matrices to model the original parameter matrix. These structured matrices can be transformed into very low rank matrices via some mathematical operations. Denton \etal~\cite{Denton14NIPS} adopt matrix factorization to explore the linear structure of neural networks. They use singular value decomposition (SVD) to construct the approximation.

\textbf{Network pruning:} Network pruning is a classic topic in model compression, which has been widely studied for a long time. In early 90s of the 20th century, pruning has already been adopted to reduce the number of connections and prevent over-fitting~\cite{LeCun90NIPS, Hassibi93NIPS}. Recently, Han \etal~\cite{Han15NIPS} proposed a pruning method to remove the redundancy of deep models. Small-weight connections below a threshold would be discarded, leading to a sparse architecture. But their method did not reduce the size of activation tensor, which would dominate the memory footprint when batch size is large.  Thus some researchers focus their attention on filter pruning to reduce channel number of activation tensor. Hu \etal~\cite{Hu16arxiv} proposed a data-driven neuron pruning approach to remove unimportant neurons. Li \etal~\cite{Li16arxiv} pruned neurons via a similar strategy. They use the absolute weight sum to measure the importance of each filter, and less useful filters are dropped. Our approach is similar to these methods, but use a completely different strategy both in channel selection and model fine-tuning.

\textbf{Parameter quantization:} Parameter quantization is another classic compression method that has been well studied. One of the widely used methods is product quantization~\cite{Jegou11TPAMI}, which decomposes the space into a Cartesian product of low-dimensional subspaces and quantizes each subspace separately. Gong \etal~\cite{Gong14arxiv} compared product quantization with several different quantization methods, and found that even with a simple k-means based method, their approach could achieve impressive results. Chen \etal~\cite{Chen15ICML} introduced the hashing trick into model compression, and proposed HashedNets. All connections mapped into the same hash bucket shared the same parameter value. Similar strategy is also adopted in the Deep Compression method~\cite{Han15NIPS}. However, ~\cite{Gong14arxiv, Han15NIPS} required the parameter matrices to be reconstructed during test phase, which limited its further practical application. In order to reduce the run-time memory consumption, Wu \etal~\cite{Wu16CVPR} proposed a novel quantization method, which achieved simultaneous acceleration and compression for both convolutional and fully-connected layers.

\textbf{Designing more compact architecture:} A typical CNN architecture can be divided into two major parts: convolutional layers and fully-connected layers. While fully-connected layers dominate the majority of all parameters, recent networks usually replace fully-connected layers with average pooling. This strategy has been widely adopted in Network-In-Network~\cite{Lin13arxiv}, GoogLeNet~\cite{Szegedy15CVPR} and ResNet~\cite{He16CVPR}. In this paper, we also use this strategy to remove fully-connected layers. On the other hand, some specially designed architectures have also been explored. For example, Iandola \etal~\cite{Iandola16arxiv} use a fundamental building block called ``Fire module'' to construct the whole network, and achieve AlexNet~\cite{Krizhevsky12NIPS} level accuracy with only 4.8MB disk size, while the original AlexNet has more than 200MB model size.

\textbf{Knowledge distillation:} Knowledge distillation can be regarded as a kind of transfer learning, which aims to transform the knowledge learned by a cumbersome model to a simple model. The original cumbersome model plays the role of a teacher, and a small model is trained to mimic the performance of the teacher. With the guidance from the teacher model, the student model can even outperform this teacher~\cite{Luo16AAAI}. Ba and Caruana~\cite{Ba14NIPS} proposed to use the logits to train the mimic model. However, in order to achieve a comparable accuracy, the mimic model must have the same number of parameters. Hinton \etal~\cite{Hinton14NIPS} used a temperature parameter to control the soft level of probability distribution, and received better results. Luo \etal~\cite{Luo16AAAI} revealed that the top hidden layer preserves much more knowledge than logits when transfer to some domain-specific tasks like face recognition.

\textbf{Binary network:} In most deep learning models, we use a 32-bit floating point number to represent a parameter. If it can be binarized, model size can be reduced $32\times$, and vector multiplication can be quickly finished with dedicated hardware like the XNOR gate. Therefore, binary network has become an important research field in model compression. Formerly, a binary network is very hard to converge even on small scale datasets such as MNIST. Recently, Courbariaux \etal~\cite{Courbariaux15NIPS} proposed BinaryConnect and achieved comparable accuracy to normal networks on these datasets. However, BinaryConnect only binarized weights, the activation of each layer is still a real-valued tensor. Thus Courbariaux \etal further binarized both weights and activation in~\cite{Courbariaux16NIPS}. At the same time, Rastegari \etal~\cite{Rastegari16ECCV} started the exploration of binary network on large scale dataset like ImageNet, and achieved AlexNet level accuracy.

\section{Entropy-based pruning method}

\begin{figure*}
  \centering
		\includegraphics[width=0.9\linewidth]{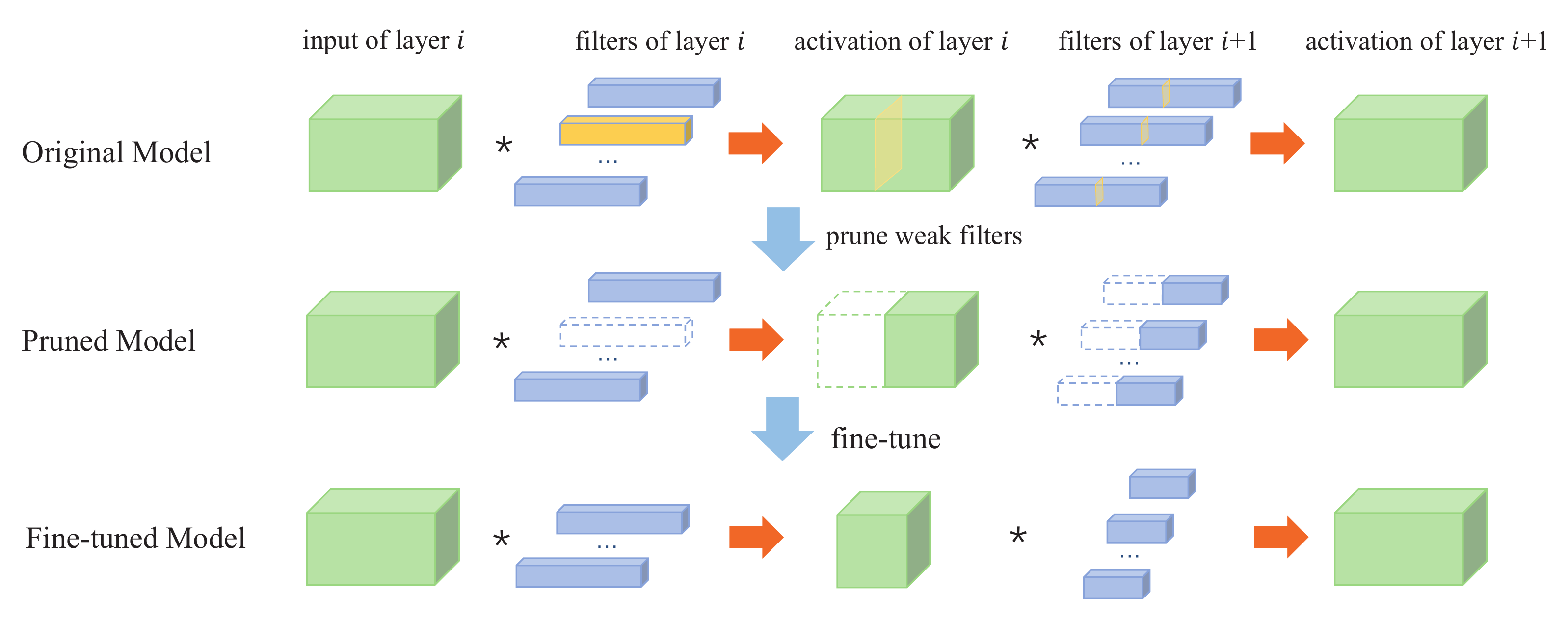}
	\caption{Illustration of our method. First, we use the entropy-based criterion to select several weak filters (highlighted in yellow on the top row). These weak filters have little contribution to the overall performance, thus can be discarded, leading to a pruned model. Finally, the whole network is fine-tuned to recover its generalization ability. $\ast$ is the convolution operator. (This figure is best viewed in color.)}
	\label{framework}
\end{figure*}

In this section, we give a comprehensive description of our intermediate activation pruning approach. First, the overall framework is presented. Our main idea is to discard several unimportant filters, and to recover its performance via fine-tuning. These implementation details would be revealed subsequently. Finally, we introduce an efficient learning schedule strategy, which is one of the key elements of our approach.

\subsection{Framework}
Figure~\ref{framework} illustrates the overall framework of our proposed intermediate activation pruning approach. For a specific layer that we want to prune (\ie, layer $i$), we first focus on its activation tensor. If several channels of its activation tensor are weak enough (\eg, all elements have the same value), we have enough confidence to believe that the corresponding filters are also less important, which could be pruned. We proposed an entropy-based metric to evaluate the weakness of each channel. As shown in Figure~\ref{framework}, these weak channels are highlighted in yellow.

Then, these weak filters are all removed from the original model, leading to a more compact network architecture. The corresponding channels of filters in the next layer are also removed. This pruned model has much fewer parameters compared with its original model, thus both running time and memory consumption can be reduced. More importantly, the size of activations are also reduced, which is less concerned in previous compression methods. 

Finally, according to the property of sparse and distributed representations (\ie, each concept is represented by many neurons, each neuron participates in the representation of many concepts~\cite{Hinton86CogSci, Bengiobeng13TPAMI}), although these filters are weak, there are also some knowledge stored in them. Thus, generalization ability of the pruned model will be affected. In order to recovery its performance, the whole network is fine-tuned. We adopt a different learning schedule to train the pruned model, which not only reduces the overall training time, but also prevents it from being attracted to bad local minima.

\subsection{Entropy-based filters selection} \label{entropy_selection}
We use a triplet $\langle \mathcal{I}_i, \mathcal{W}_i, \ast\rangle$ to denote the convolution in layer $i$, where $\mathcal{I}_i \in \mathbb{R}^{c\times h \times w}$ is the input tensor, $\mathcal{W}_i \in \mathbb{R}^{d\times c \times k \times k}$ is a set of filter weights, $\ast$ denote the convolution operation. Our goal is to prune the filters $\mathcal{W}_i$.

Note that each filter corresponds to a single channel of its activation tensor $\mathcal{I}_{i+1}$ (the activation of layer $i$ and at the same time input of layer $i+1$, shown as the middle green block in Figure~\ref{framework}), the discriminative ability of each filter is closely related to its activation channel. A simple strategy is to calculate the average percentage of zeros (APoZ) in each activation channel~\cite{Hu16arxiv}. But this is not an optimal metric. First, a small value (\eg, 0.0001, close to but not exactly zero) would be omitted, which should be pruned in fact. Secondly, if a filter always produces similar values, we can believe that this filter contains less information, thus is less important. However, it is impossible to select this filter with the APoZ metric.

In this paper, we propose an entropy-based metric to evaluate the importance of each filter. Entropy is a commonly used metric to measure the disorder or uncertainty in information theory. A larger entropy value means the system contains more information. In our filter pruning scenario, if a channel of activation tensor contains less information, its corresponding filter is less important, thus could be dropped.

We first use global average pooling to convert the output of layer $i$, which is a $c\times h\times w$ tensor, into a $1\times c$ vector. In this way, each channel of $\mathcal{I}_{i+1}$ (activation of layer $i$ / input of layer $i+1$) has a corresponding score for one image. In order to calculate the entropy, more output values need to be collected, which can be obtained using an evaluation set. In practice, the evaluation set can be simply the original training set, or a subset of it. Finally, we get a matrix $\mathcal{M} \in \mathbb{R}^{n\times c}$, where $n$ is the number of images in the evaluation set, and $c$ is the channel number. For each channel $j$, we would pay attention to the distribution of $\mathcal{M}_{:,j}$. To compute the entropy value of this channel, we first divide it into $m$ different bins, and calculate the probability of each bin. Finally, the entropy can be calculated as follows:
\begin{equation}
	H_j=-\sum_{i=1}^m p_i\log p_i.
\end{equation}
Where, $p_i$ is the probability of bin $i$, $H_j$ is the entropy of channel $j$. In general, if some layers are weak enough, \eg, most of their activation are zeros, their entropy are relatively small. Hence, our entropy-based method can be used for evaluating the importance of each channel. A smaller score of $H_j$ means channel $j$ is less important in this layer, thus could be removed.

The next issue is how to decide the pruning boundary. One feasible method is to specify a threshold value, all channels with score below this threshold are removed from the network. However, this threshold value is a hyperparameter, which is hard to be specified. Another more practical method is using a constant compression rate. All the filters are sorted in the descending order according to their entropy scores, and only the top $k$ filters are preserved. Of course, the corresponding channels in $\mathcal{W}_{i+1}$ are removed too.

\begin{table*}[!t]
	\caption{Overall performance of our approach to reduce FLOPs\protect\footnotemark[2] and parameters on the VGG-16 model. Note that the total activation size is the sum of convolutional, relu, pooling layers' output and the input data when batch size is set to 1.}
	\label{params}
	\setlength{\tabcolsep}{4pt} 
	\centering
		\begin{tabular}{c||c|c|c||c|c|c||c|c|c}
			\hline
			\multirow{2}{*}{Layer} & \multicolumn{3}{c||}{\#FLOPs}  & \multicolumn{3}{c||}{\#Parameters}   & \multicolumn{3}{c}{Intermediate Activation Size}\\
			\cline{2-10} & Original & Pruned & Percentage & Original & Pruned & Percentage & Original & Pruned & Percentage\\
			\hline
			Conv1-1 & 86.7M & 43.35M & 50\% & 1.73K & 0.86K & 50\% & 12.25MB & 6.125MB & 50\% \\
			Conv1-2 & 1.85B & 462.42M & 25\% & 36.86K & 9.22K & 25\% & 12.25MB & 6.125MB & 50\% \\
			\hline
			Conv2-1 & 0.92B & 231.21M & 25\% & 73.73K & 18.43K & 25\%  & 6.13MB & 3.06MB & 50\% \\
			Conv2-2 & 1.85B & 462.42M & 25\% & 147.46K & 36.86K & 25\% & 6.13MB & 3.06MB & 50\% \\
			\hline
			Conv3-1 & 0.92B & 231.21M & 25\% & 294.91K & 73.73K & 25\% & 3.06MB & 1.53MB & 50\% \\
			Conv3-2 & 1.85B & 462.42M & 25\% & 589.82K & 147.46K & 25\% & 3.06MB & 1.53MB & 50\% \\
			Conv3-3 & 1.85B & 462.42M & 25\% & 589.82K & 147.46K & 25\% & 3.06MB & 1.53MB & 50\% \\
			\hline
			Conv4-1 & 0.92B & 231.21M & 25\% & 1.18M & 294.92K & 25\% & 1.53MB & 0.77MB & 50\% \\
			Conv4-2 & 1.85B & 462.42M & 25\% & 2.36M & 589.82K & 25\% & 1.53MB & 0.77MB & 50\% \\
			Conv4-3 & 1.85B & 462.42M & 25\% & 2.36M & 589.82K & 25\% & 1.53MB & 0.77MB & 50\% \\
			\hline
			Conv5-1 & 462.42M & 231.21M & 50\% & 2.36M & 1.18M & 50\% & 392KB & 392KB & 100\% \\
			Conv5-2 & 462.42M & 462.42M & 100\% & 2.36M & 2.36M & 100\%& 392KB & 392KB & 100\% \\
			Conv5-3 & 462.42M & 462.42M & 100\% & 2.36M & 2.36M & 100\% & 392KB & 392KB & 100\% \\
			\hline
			FC6 & 102.76M & -- & -- & 102.76M & -- & -- & 16KB & -- & -- \\
			FC7 & 16.78M & -- & -- & 16.78M & -- & -- & 16KB & -- & -- \\
			FC8 & 4.10M & 512K & 12.48\% & 4.10M & 512K & 12.48\% & 3.91KB & 3.91K & 100\% \\
			\hline
			Total & 15.5B & 4.67B & 30.32\% & 138.34M & 8.32M & 6.01\% & 109.89MB & 56.33MB & 51.26\%\\
			\hline
		\end{tabular}
\end{table*}

\subsection{Pruning strategy} \label{mylabel1}% vgg and resnet
There are mainly two kinds of different network architectures: the traditional convolutional and fully-connected architecture, and some structure variants. The former is represented by Alex-Net~\cite{Krizhevsky12NIPS} or VGG-Net~\cite{Simonyan14ICLR}, while the latter mainly include some recent networks like GoogLeNet~\cite{Szegedy15CVPR} and ResNet~\cite{He16CVPR}. The main difference between these two kinds of architectures is that more recent networks usually replace the fully-connected layers with a global average pooling layer, and adopt some new network structures like Inception in GoogLeNet and residual blocks in ResNet.

We use two different strategies to prune these two types of networks. For VGG-16,  we notice that more than 90\% FLOPs exist in the first 10 layers (Conv1-1 to Conv4-3), and the fully-connected layers contribute nearly 89.36\% of all model parameters. In order to reduce the running time, we prune the first 10 convolutional layers via our entropy-based method. As for fully-connected layers, our method is also valid, but we think replacing them with a global average pooling layer is more efficient, since our goal is to reduce the parameter size as far as possible.

\begin{figure}[t]
	\begin{center}
		\includegraphics[width=1.0\linewidth]{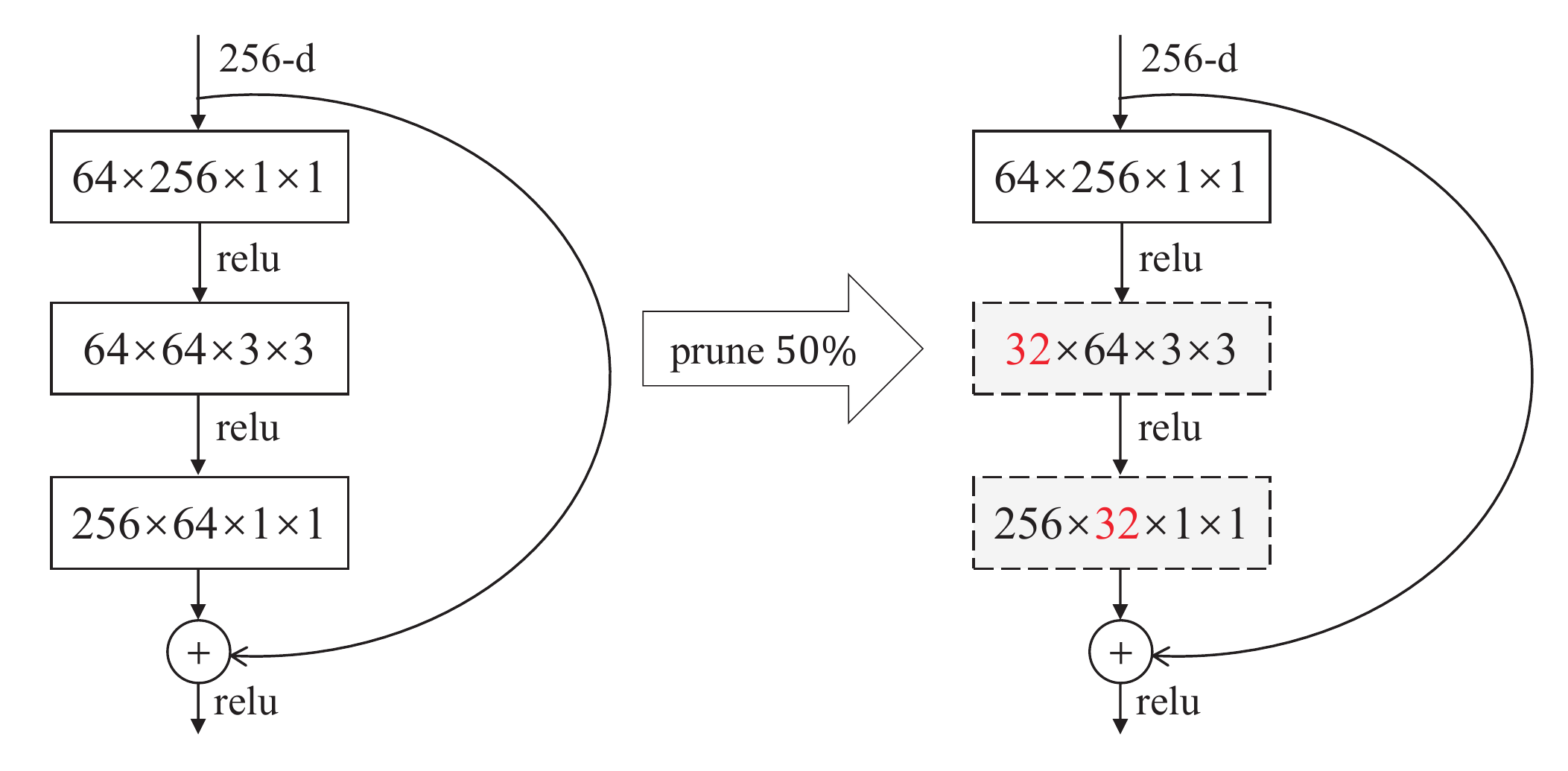}
	\end{center}
	\caption{Illustration of our ResNet pruning strategy. For each residual block, we only prune the middle convolutional layer, which would simultaneously reduce the parameters of the next layer.}
	\label{ResNet}
\end{figure}

For ResNet, there exist many restrictions due to its special structure. For example, the output channel number of each block in the same group needs to be consistent, otherwise, a special projection shortcut would be necessary in order to finish the sum operation (see~\cite{He16CVPR} for more details). Thus it is hard to prune the last convolutional layer of each residual block directly. Since most parameters are located in the last two layers, we think pruning the intermediate layer is a much better choice, which would reduce the parameter size of the next layer too. This strategy is illustrated in Figure~\ref{ResNet}.

\footnotetext[2]{FLOPs: FLoating-point OPerations. Here we regard the vector multiplication as a float-point operation as described in~\cite{He16CVPR}. Note that some papers regard this operation as two floating-point operations (multiplication and addition), thus their results are twice as large as ours.}
\subsection{Learning schedule} \label{l3}
So far, we have introduced our entropy-based channel selection method and pruning strategy for a specific layer. The remaining issue is how to prune the whole network. As revealed in~\cite{Li16arxiv}, there are two strategies to solve this problem: \textit{prune once and retrain} or \textit{prune and retrain iteratively}. We demonstrate in our experiments that the former strategy is not feasible since pruning too many layers may drop the accuracy significantly. Once the performance is heavily damaged, it cannot be regained by fine-tuning the pruned model. The latter strategy, however, requires too many epochs to fine-tune the whole network, especially when the network is very deep.

In this paper, we propose a novel learning schedule to trade-off between training speed and accuracy. Whenever one layer is pruned, the whole network is fine-tuned with one or two epochs to recover its performance slightly. Of course, this performance is not yet optimal, but is approaching to it closely. Only after the final layer has been pruned, the network is fine-tuned carefully with many epochs. Using such a learning schedule, the overall training time can be reduced dramatically. But more importantly, this strategy would prevent the fine-tuned network from sinking into bad local minima in the early learning stage. Because the training process of deep neural networks is a highly non-convex, if we always fine-tuned the network until convergence after pruning every layer, the network is likely to be attracted to a set of poor values in the early stage.

\section{Experiments}
We pruned two typical CNN models using our pruning approach: VGG-16 and ResNet-50, which represent two different CNN architectures respectively. Both models are fine-tuned on the ImageNet dataset (ILSVRC-12~\cite{ILSVRC15}) using the standard Caffe library~\cite{jia2014caffe}. The ImageNet dataset consists of over one million training images drawn from 1000 categories. We random select 10 images from all 1000 categories in the training set to comprise our evaluation set (used for channel selection in section~\ref{entropy_selection}). Top-1 and top-5 classification performance are reported on the 50k validation set, using the single-view testing approach (central patch only). All experiments are run on a computer equipped with Nvidia Tesla K80 GPU.

\subsection{VGG-16 on ImageNet} \label{l4}
As described previously, we prune the network from Conv1-1 to Conv4-3 iteratively to accelerate both training and inference speed, and discard the fully-connected layer to reduce the parameter size. During training, all images are resize to $256\times256$, then $224\times 224$ random crop is adopted to feed the data into the network. Horizontal flip is also used for data augmentation. At the inference stage, we center crop the resized images to $224\times 224$. Other augmentation tricks are not used in our experiments. Except for the last layer (Conv4-3), the network is fine-tuned quickly in one or two epochs with $10^{-3}$ or $10^{-4}$ learning rate. During the last layer pruning, we fine-tune the whole network carefully in 12 epochs with learning rate varying from $10^{-3}$ to $10^{-5}$. We use SGD with mini-batch size of 128, and other parameters are kept the same as the original VGG paper~\cite{Simonyan14ICLR}.

The detailed distribution of FLOPs and parameters in each layer is shown in Table~\ref{params}. As we can see, the first 10 layers contribute more than 90\% FLOPs. Similar phenomenon is also appeared in the intermediate activation size. We prune each layer with 50\% compression rate, \ie, discarding half of the filters. With such a strategy, the pruned model gains $3.3\times$ speed-up both in training and inference stages. In practice, we test these two models with a mini-batch size of 50 on a K80 GPU. As expected, the original VGG-16 model takes 863.04ms/2268.74ms during forward/backward stage, while the pruned model only takes 349.21ms/869.45ms respectively. 

In order to reduce parameter size, we focus on fully-connected layers. Although our approach is also suitable for FC layers, we think replacing them with a global average pooling layer is much better. First, the existence of fully-connected layers seems more likely to be overfitting. As revealed in~\cite{Lin13arxiv}, the global average pooling layer summarizes the spatial information via average values, thus can be seen as a regularizer. Secondly, many previous works~\cite{Lin13arxiv, Szegedy15CVPR} have demonstrated that global average pooling can even outperform FC layers with the same network architecture.

\begin{table}
	\caption{Performance changes of the VGG-16 model before/after pruning with the proposed approach.}
	\label{vgg16}
	\setlength{\tabcolsep}{3pt} 
	\centering
		\begin{tabular}{c|c|c|c|c}
			\hline
			Model & Top-1 & Top-5 & Speed-up & Compression\\
			\hline
			Original\protect\footnotemark[3]  & 68.36\% & 88.44\% & $1\times$ & $1\times$\\
			Pruned-FC & 69.12\% & 88.89\% & $3.24\times$ & $1.05\times$ \\
			train from scratch & 67.00\% & 87.45\% & $3.24\times$ & $1.05\times$ \\
			Pruned-GAP & 66.80\% & 87.28\% & $3.30\times$ & $16.64\times$ \\
			\hline
		\end{tabular}
\end{table}
\footnotetext[3]{For a fair comparison, the accuracy of original VGG-16 model is evaluated on resized center-cropped images using pre-trained model as adopted in~\cite{Han15NIPS, Hu16arxiv}. The same strategy is also used in ResNet-50.}

We summarized the performance of the proposed approach in Table~\ref{vgg16}. ``Pruned-FC'' refers to the model in which only the first 10 convolutional layers are pruned for speed consideration. Because some useless filters are discarded, the pruned model can even outperform the original model. But if we train the same structure as ``Pruned-FC'' from scratch, the accuracy could be much lower. This phenomenon is reasonable, since weight initialization plays a crucial role in CNN training. We then remove the FC layers, replaced with a global average pooling layer, and the final pruned model (``Pruned-GAP'') is fine-tuned to regain its accuracy. The final classification accuracy of our pruned model is slightly lower than original model, since the model size has been reduced greatly. Further pruning could also been adopted if somebody is more concerned about getting a small scale model.

Finally, we compare our method with following baselines on the VGG-16 model:
\begin{itemize}
	\item \textbf{Pruning}~\cite{Han15NIPS}: An iteratively pruning methods, which discard small-weight connections and fine-tune to recover its performance, leading to a sparse network.
	\item \textbf{APoZ}~\cite{Hu16arxiv}: Similar to our method, but use different channel selection metric and training strategy. They calculate the sparsity of each channel in output activations as its importance score.
	\item \textbf{Taylor expansion}~\cite{Taylor}: This criterion believes a filter can be safely removed if it has little influence on the loss function. Thus they use Taylor expansion to approximate the loss change.
\end{itemize}

These methods all achieve good results when convert a cumbersome VGG model into a simpler model. As shown in Table~\ref{vgg16_compare}, our method generates comparable performance with them. 

\begin{table}
	\caption{Comparison of different model compression methods on VGG-16 network.}
	\label{vgg16_compare}
	\small
	\setlength{\tabcolsep}{2.5pt} 
	\centering
		\begin{tabular}{c|c|c|c|c}
			\hline
			Method &Top-1 Acc.&Top-5 Acc.&Speed-up&Compression\\
			\hline
			Pruning~\cite{Han15NIPS} & +0.16\% &+0.44\% & $5\times$ & $13\times$ \\
			APoZ-1~\cite{Hu16arxiv} & -2.16\% & -0.84\% & -- & $2.04\times$ \\
			APoZ-2~\cite{Hu16arxiv} & +1.81\% & +1.25\% & -- & $2.70\times$ \\ 
			Taylor-1~\cite{Taylor} & -- & -1.44\% & $2.68\times$ & -- \\
			Taylor-2~\cite{Taylor} & -- & -3.94\% & $3.86\times$ & -- \\
			\hline
			Pruned-FC & +0.76\% & +0.45\% & $3.24\times$ & $1.05\times$ \\
			Pruned-GAP & -1.56\% & -1.16\% & $3.30\times$ & $16.64\times$ \\
			\hline
		\end{tabular}
\end{table}

Among these methods, the ``Pruning'' approach proposed by Han~\etal~\cite{Han15NIPS} achieves pretty impressive performance. Yet in spite of this, their non-structured sparse model can not be supported by any off-the-shelf deep learning libraries. Hence, some specialized hardwares and softwares are needed for efficient inference, which is difficult and expensive in real-world applications.

APoZ~\cite{Hu16arxiv} and Taylor expansion-based method~\cite{Taylor} are similar to our framework but using difference selection criterion. However, they performs much worse than ours. APoZ-1 only pruned few layers (Conv4, Conv5 and the FC layers), but leads to accuracy degradation. APoZ-2 pruned Conv5-3 and the FC layers. The accuracy is slightly improved but this model has little influence on acceleration the training or testing process. 

In contrast, Molchanov~\etal~\cite{Taylor} focus on model acceleration and only prune the convolutional layers. Such a strategy is also adopted in ``Pruned-FC''. As shown in Table~\ref{vgg16_compare}, the Taylor method would lead to a significant performance degradation, while our method can even improve the model accuracy.

\subsection{ResNet-50 on ImageNet}\label{l2}
The proposed pruning method is also effective to compress recent CNN variants. We explore model compression on the ResNet-50~\cite{He16CVPR} architecture, which achieves state-of-the-art classification accuracy on ImageNet. For convenience, We only prune the middle convolutional layer as we introduced in Section~\ref{mylabel1}. 

The filters are pruned iteratively from block 2a to 5c. Except for the last block, the pruned network is only fine-tuned one epoch with learning rate of $10^{-4}$ to recover its performance. As for block 5c, we fine-tune it carefully with 4 epochs, varying the learning rate from $10^{-4}$ to $10^{-5}$. Other parameters are kept the same as our VGG-16 pruning.

\begin{table}
	\caption{Overall Performance of our approach on ResNet-50 with different compression rate.}
	\label{resnet50}
	\footnotesize
	\setlength{\tabcolsep}{3.5pt}
	\centering
		\begin{tabular}{c|c|c|c|c|c|c}
			\hline
			\multirow{2}{*}{Model} & \multirow{2}{*}{Top-1} & \multirow{2}{*}{Top-5} & \multicolumn{2}{c|}{Speed-up}  & \multicolumn{2}{c}{Compression}\\
			\cline{4-7} & & & \#FLOPs & FLOPs\% & \#Para. & Para.\%\\
			\hline
			Original & 72.88\% & 91.14\% & 3.86B & 100\% & 25.56M & 100\% \\
			Pruned-90 & 73.56\% & 91.60\% & 3.58B & 93\% & 23.89M & 93\%  \\
			Pruned-75 & 72.89\% & 91.27\% & 3.19B & 83\% & 21.47M & 84\%  \\
			Pruned-50 & 70.84\% & 90.03\% & 2.52B & 65\% & 17.38M & 68\%  \\
			\hline
		\end{tabular}
\end{table}

The overall performance of our method on pruning ResNet-50 is shown in Table~\ref{resnet50}. We prune this model with 3 different compression rate (preserve 90\%, 75\%, 50\% filters respectively). Unlike traditional CNN architecture, the ResNet is much more compact. There exists less redundancy compared with the VGG-16 model, thus pruning a large amount of filters seems more difficult. In spite of this, our method can even improve the performance of ResNet-50 when a small compression rate is adopted.

Since ResNet is a recent proposed model, the literature lacks enough works aiming to compress this network. Thus, we could only compare our method with others for a rough comparison. In~\cite{Li16arxiv}, Li \etal proposed a similar framework for filter pruning. They calculated the absolute weight sum of each filter as the channel selection metric. They pruned ResNet-34 on ImageNet to get a smaller model. However, their method can not prune too many filters. Otherwise, the generalization ability of pruned model would be damaged greatly. In~\cite{Li16arxiv}, they only pruned 10.8\% parameters with 1.06\% drop in the top-1 accuracy. As a comparison, our model ``Pruned-75'' pruned 16\% parameters and slightly increased both top-1 and top-5 accuracy rates.

\subsection{Effectiveness analysis}
We now demonstrate the effectiveness of our learning schedule, which gives a good trade-off between training speed and accuracy.

We first compare two commonly used strategies in network pruning: \textit{prune once} and \textit{prune iteratively}, as we introduced in Section~\ref{l3}. Since our learning schedule is based on the iterative pruning strategy, we use the performance of our method to represent the former strategy. As shown in Table~\ref{vgg16}, the Pruned-FC model achieves $3.24\times$ speed-up as well as accuracy improvement. Then, we prune the same layers at one time, and fine-tune it in the same number of epochs. In order to avoid the interference of former pruning, we prune these layers from Conv4-3 to Conv1-1. In this case, the output activations for entropy computing are not changed. Unfortunately, the final Top-1/Top-5 accuracy is only 58.82\%/81.91\%, much worse than the iterative strategy.  

This result is reasonable, since pruning too much filters at one time would greatly harm the overall generalization ability of CNN model, thus is hard to regain its original performance. In fact, if we adopt the same network architecture, but train the model from scratch, we can not get a same level accuracy.

But if we always fine-tune the pruned model until accuracy is recovered, the time cost would be intolerable especially when the network is very deep. Thus we propose a learning schedule to balance training speed and accuracy. These two methods are compared in Table~\ref{t6}.

\begin{table}
	\caption{Comparison of our learning schedule strategy and a naive iterative pruning method.}
	\label{t6}
	\setlength{\tabcolsep}{3pt} 
	\centering
		\begin{tabular}{c|c|c|c|c}
			\hline
			Method & Layer & epochs & Top-1 & Top-5 \\
			\hline
			\multirow{3}{*}{Fine-tune Carefully} & Conv5-3 & 12 & 70.49\% & 89.99\% \\
			& Conv5-2 & 12 & 70.33\% & 89.74\% \\
			& Conv5-1 & 12 & 69.81\% & 89.54\% \\
			\hline
			\multirow{3}{*}{Learning Schedule} & Conv5-3 & 1 & 64.85\% & 86.87\% \\
			& Conv5-2 & 1 & 64.69\% & 86.66\% \\
			& Conv5-1 & 12 & 70.09\% & 89.76\% \\
			\hline
		\end{tabular}
\end{table}

Here, we prune the Conv5 layer with a reverse order (\ie, from Conv5-3 to Conv5-1), and report the final accuracy of each layer. As we can see, the naive iterative pruning method is extremely time-consuming. It can take weeks or even months for model training. But more importantly, this strategy can easily lead the model into a local minima. However, our learning schedule strategy is more efficient. It reduces training epochs dramatically, but generates better classification accuracy, achieving a great balance between high accuracy and fast train speed.

\subsection{Domain adaptation of the pruned model}
One of the main differences between our pruning method and other current model compression methods is that our pruned model can be easily transfered into other domains. In this experiment, we will evaluate the domain adaptation performance of the pruned model on two different recognition tasks: fine-grained classification and scene recognition.

We compare the pruned VGG-16 network with the original VGG-16 on two classic benchmark datasets:
\begin{itemize}
	\item \textbf{CUB-200-2011}~\cite{CUB_200_2011}: A typical dataset for fine-grained classification task, which aims to recognize bird sub-categories. The CUB-200-2011 dataset contains 11,788 images of 200 different bird species. We only use the raw image data and corresponding labels to fine-tune CNN models. No other supervised information (\eg, bounding box) is used for simplicity. We replace the softmax layer with 200 category output, and fine-tune the new network in 30 epochs with batch size of 30, varying learning rate from $10^{-3}$ to $10^{-4}$.
	\item \textbf{Indoor-67}~\cite{Indoor67}: This dataset contains 67 categories for indoor scene recognition, which is a challenging open problem in high level vision task. We follow the official splitting to organize the training/validation set, containing 5360/1340 images respectively. The processed network is fine-tuned in 20 epochs with batch size of 32, varying learning rate from $10^{-3}$ to $10^{-5}$.
\end{itemize}

\begin{table}
	\caption{Performance comparison of our pruned model and the original VGG-16 network on different benchmark dataset.}
	\label{t7}
	\setlength{\tabcolsep}{2.5pt} 
	\centering
		\begin{tabular}{c|c|c|c}
			\hline
			Dataset & Model & Top-1 & Top-5 \\
			\hline
			\multirow{3}{*}{CUB-200-2011~\cite{CUB_200_2011}} & Original VGG-16 & 71.95\% & 90.77\% \\
			& Pruned-FC & 69.55\% & 89.78\% \\
			& Pruned-GAP & 69.69\% & 89.54\% \\
			\hline
			\multirow{3}{*}{Indoor-67~\cite{Indoor67}} & Original VGG-16 & 72.16\% & 93.51\% \\
			& Pruned-FC & 70.07\% & 92.31\% \\
			& Pruned-GAP & 69.10\% & 91.72\% \\
			\hline
		\end{tabular}
\end{table}

The overall performance is summarized in Table~\ref{t7}. An additional model named \textit{Pruned-FC} (\ie, only convolutional layers till Conv4-3 are pruned) is added for comparing the influence of fully-connected layers for domain adaptation.

From Table~\ref{t7}, we can discover that the classification accuracy of our pruned model decreases slightly compared with the original VGG-16 model, but still in an acceptable range. In the CUB-200-2011 dataset, we find that {Pruned-GAP} achieves acceptable or even better performance than {Pruned-FC}. This phenomenon demonstrates that fully-connected layers are heavily redundant, which can even result in over-fitting. But fine-grained recognition is a challenging task, it may need more channels to represent the subtle  inter-class difference. Thus, both pruned model is slightly worse than the original model.

As for Indoor-67, the similarity between ImageNet and this dataset is relatively lower than others. When transfer to this domain, fully-connected layers can help the network to build a stronger classifier than a single softmax. Since the parameters and FLOPs of our pruned model have been dramatically reduced, we think this accuracy degradation is acceptable.

\section{Conclusion}
In this paper, we proposed an entropy-based framework to simultaneously accelerate and compress CNN models in both training and inference stages. The pruned model shows better performance compared with previous pruning strategies. Our method does not rely on any dedicated library, thus can be widely used in various practical applications with current deep learning libraries.

In the future, we would like to design a dynamic pruning framework. An alternative method for better channel selection is also worthy to be studied. Extensive exploration on more vision tasks (such as object detection, semantic segmentation and depth estimation) with the pruned networks is an interesting direction too. The pruned networks will greatly accelerate these vision tasks.

\clearpage
{\small
\bibliographystyle{ieee}
\bibliography{mybib}
}

\end{document}